\documentclass[10pt,twocolumn,letterpaper]{article}

\usepackage{iccv}
\usepackage{times}
\usepackage{epsfig}
\usepackage{graphicx}
\usepackage{amsmath}
\usepackage{amssymb}

\usepackage{glossaries}
\usepackage{microtype} 
\usepackage{subfigure}

\usepackage[breaklinks=true,bookmarks=false]{hyperref}

\iccvfinalcopy 

\def\iccvPaperID{19} 

\ificcvfinal\fi

\newacronym{aes}{AES}{Advanced Encryption Standard}

\newacronym{api}{API}{application programming interface}

\newacronym{cpu}{CPU}{Central Processing Unit}

\newacronym{gpu}{GPU}{Graphics Processing Unit}

\newacronym{mse}{MSE}{mean squared error}

\newacronym{otp}{OTP}{one-time pad}

\begin{document}

\title{
On the Importance of Encrypting Deep Features
}

\author{
Xingyang Ni\\
Tampere University\\
Tampere, Finland\\
{\tt\small xingyang.ni@tuni.fi}
\and
Heikki Huttunen\\
Visy Oy\\
Tampere, Finland\\
{\tt\small heikki.huttunen@visy.fi}
\and
Esa Rahtu\\
Tampere University\\
Tampere, Finland\\
{\tt\small esa.rahtu@tuni.fi}
}

\maketitle
\ificcvfinal\thispagestyle{empty}\fi

\begin{abstract}

In this study, we analyze model inversion attacks with only two assumptions: feature vectors of user data are known, and a black-box \acrshort{api} for inference is provided.
On the one hand, limitations of existing studies are addressed by opting for a more practical setting.
Experiments have been conducted on state-of-the-art models in person re-identification, and two attack scenarios (\ie, recognizing auxiliary attributes and reconstructing user data) are investigated.
Results show that an adversary could successfully infer sensitive information even under severe constraints.
On the other hand, it is advisable to encrypt feature vectors, especially for a machine learning model in production.
As an alternative to traditional encryption methods such as \acrshort{aes}, a simple yet effective method termed ShuffleBits is presented.
More specifically, the binary sequence of each floating-point number gets shuffled.
Deployed using the \acrlong{otp} scheme, it serves as a plug-and-play module that is applicable to any neural network, and the resulting model directly outputs deep features in encrypted form.
Source code is publicly available at \url{https://github.com/nixingyang/ShuffleBits}.

\end{abstract}

\section{Introduction}

Due to the availability of large-scale datasets~\cite{zheng2015scalable,ristani2016performance,wei2018person} and affordable computing resources, the field of machine learning has witnessed rapid progress over the past decade.
Real-world applications can be found in everyday life, \eg, targeted advertising in online shopping, recommender systems in video streaming services, and virtual assistants on smart devices.
With the widespread adoption of techniques such as person re-identification~\cite{zhong2017random,lin2019improving,luo2019bag,ye2020deep,he2020fastreid}, the concern over security issues can not be overemphasized.
Significant efforts have been put into understanding the vulnerabilities in machine learning models~\cite{kurakin2016adversarial,shokri2017membership,tramer2016stealing,wu2016methodology}.
In the following, we outline four types of attacks, \ie, adversarial example attacks, membership inference attacks, model extraction attacks, and model inversion attacks.

In adversarial example attacks, input data is slightly manipulated so that a human may not observe the changes while the model would make incorrect predictions~\cite{kurakin2016adversarial}.
In~\cite{dong2018boosting}, a momentum term is integrated into the iterative process for performing attacks, and it stabilizes the direction for updates, avoids poor local maxima, and improves the success rate.
Afterward, Su \etal~\cite{su2019one} analyses an extreme case where only one pixel can be modified.
Perturbation is encoded into an array, and the candidate solution is optimized by adopting differential evolution.
By contrast, He \etal~\cite{he2017adversarial} generates an ensemble of weak defenses, while the resulting method does not always promote resilience.

In membership inference attacks, an adversary is interested in identifying whether a specific sample is included in a model's training set~\cite{shokri2017membership}.
Multiple shadow models are trained to simulate the target model while the membership in their training sets is available~\cite{shokri2017membership,long2017towards}.
Subsequently, a separate threat model is trained on the input-output pairs of the shadow models, and it behaves differently depending on whether the sample is used for training the target model.
In~\cite{yeom2018privacy}, the relation between overfitting and membership vulnerability has been studied, and results indicate that overfitting is a sufficient but not necessary condition for membership vulnerability.

In model extraction attacks, an adversary has black-box access to a target model, and the primary purpose is to duplicate the functionality of the target model~\cite{tramer2016stealing}.
Experiments on simple target models show that one could train substitute models locally on public datasets with near-perfect fidelity~\cite{tramer2016stealing}.
Under similar settings, a reinforcement learning approach is proposed in~\cite{orekondy2019knockoff} to improve sample efficiency of queries, and a real-world image recognition model was pirated with reasonable performance.
Juuti \etal~\cite{juuti2019prada} design a countermeasure that analyses the distribution of consecutive query requests and raises the alarm when suspicious activities are detected.
Later on, two defense strategies are presented in~\cite{krishna2019thieves}: the first membership inference strategy checks whether inputs are outliers, and the second watermarking strategy generates wrong outputs deliberately for a tiny fraction of queries.

In model inversion attacks, an adversary intends to infer input data from a released model~\cite{wu2016methodology}.
Fredrikson \etal~\cite{fredrikson2014privacy} managed to invert a linear regression model and predict the patient's genetic markers based on demographic information.
With confidence scores returned by a facial recognition model, one could recover face images that are representative of a specific person in the training set~\cite{fredrikson2015model}.
In the case that a partial prediction vector is returned, truncation is applied to feature vectors when training the inversion model in~\cite{yang2019adversarial}.
By contrast, Zhang \etal~\cite{zhang2020secret} shifts the focus to a white-box setting and theoretically proves that the vulnerability to model inversion attacks is unavoidable for models with high predictive power.

Existing studies on model inversion attacks are subject to the following limitations:
(1) The threat model is trained on the same dataset as the proprietary model~\cite{dosovitskiy2016generating,dosovitskiy2016inverting,mahendran2016visualizing,oord2016conditional};
(2) The adversary has white-box access to the proprietary model~\cite{yin2020dreaming,zhang2020secret};
(3) Experiments are limited to small-scale low-resolution datasets~\cite{fredrikson2015model,oord2016conditional,yang2019adversarial,zhang2020secret}.
To handle these problems, we investigate model inversion attacks in a more practical setting:
(1) The proprietary dataset is unavailable, and the adversary has to collect and utilize a different local dataset;
(2) A black-box \acrshort{api} for inference is provided, while the architecture and parameters of the proprietary model are unknown;
(3) Experiments are conducted on large-scale high-resolution datasets with state-of-the-art proprietary models.

In this study, our contribution is twofold:
\begin{itemize}
\itemsep0em
\item
We adopt an experimental setting that is more practical than previous works.
Only two assumptions are made, and they hold true in most, if not all, image retrieval systems.
On the one hand, an adversary has illegitimate access to feature vectors of user data.
On the other hand, the adversary can extract feature vectors of samples in a local dataset via a black-box \acrshort{api}.
Furthermore, two attack scenarios have been validated on state-of-the-art person re-identification models.
In the presence of severe constraints, results indicate that it is still feasible to recognize auxiliary attributes with decent accuracy and reconstruct user data that are recognizable.
\item
In light of the aforementioned results, we suggest that practitioners incorporate an encryption method when transferring and storing deep features.
Note that \acrfull{aes}~\cite{daemen1999aes} is the de facto standard for symmetric-key algorithms.
We introduce an alternative method termed ShuffleBits, in which the binary sequence of each floating-point number gets shuffled.
Unlike traditional encryption methods, it can be implemented as a plug-and-play module inside neural networks, and the resulting model generates encrypted deep features in a straightforward manner.
\end{itemize}

\begin{figure}[t]
\begin{center}
\includegraphics[width=1.0\linewidth]{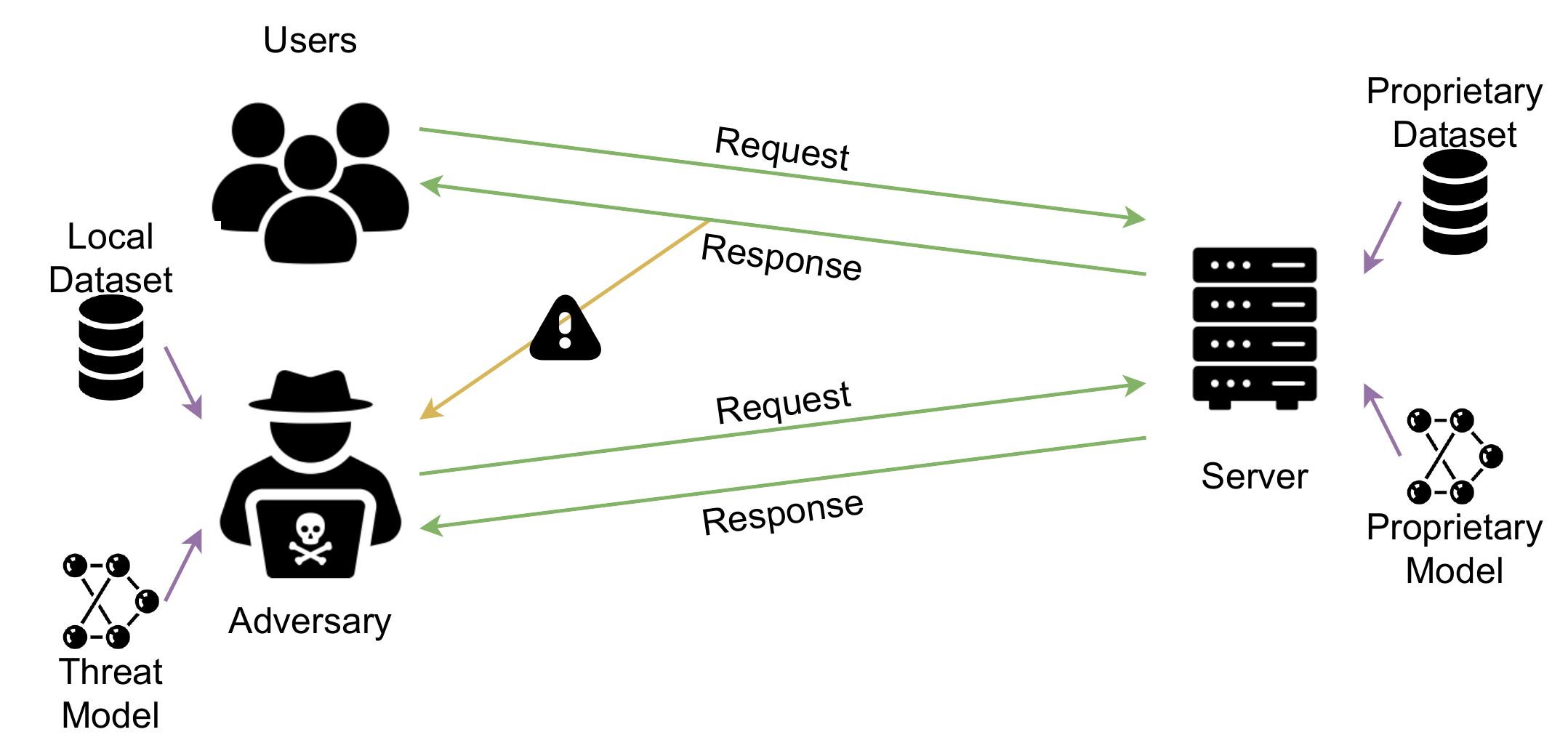}
\end{center}
\caption{
Both the users and the adversary have access to the server, in which a proprietary model is deployed.
Meanwhile, the adversary intercepts the server's responses to the users, and a local dataset is available.
}
\label{figure:attack_scenarios}
\end{figure}

\section{Attack scenarios}
\label{section:attack_scenarios}

\noindent\textbf{Preliminaries.}
Figure~\ref{figure:attack_scenarios} illustrates the background of attack scenarios in this study.
A server runs a proprietary model that is trained on a proprietary dataset, and a response containing feature vectors is returned after processing a request containing user data.
Additionally, the server's responses to the users are intercepted by the adversary, \ie, feature vectors of user data are known.
Since the proprietary dataset is unreachable, the adversary collects and utilizes a local dataset instead.
The primary purpose is to train a threat model that sniffs sensitive information of user data from the feature vectors.

\noindent\textbf{Recognizing auxiliary attributes.}
Depending on the proprietary model in question, certain auxiliary attributes may be relevant.
For example, one might be interested in a person's age and gender when using a facial recognition model.
Although the original task (\ie, recognizing faces) is inherently different from the auxiliary task (\ie, predicting age and gender), the feature vectors for the original task may still contain relevant information for solving the auxiliary task.
With a local dataset at hand, the adversary could annotate auxiliary attributes and construct a predictive model.
The multi-layer perceptron is suitable for solving multi-class classification problems, where each sample might be associated with multiple labels.

\noindent\textbf{Reconstructing user data.}
Alternatively, one could interpret the whole system as an autoencoder.
The proprietary model on the server is the encoder that maps raw data into feature vectors.
The adversary builds a decoder that reconstructs raw data from feature vectors.
The decoder is trained in an unsupervised manner, \ie, it does not require a labeled dataset.
The inputs are feature vectors extracted by the proprietary model, and the ground truth outputs are raw data.
The decoder is optimized with an objective function so that the difference between ground truth data and reconstructed data is minimized.

\newpage
\noindent\textbf{Constraints.}
Multiple constraints complicate matters for the adversary.
Firstly, the proprietary model and the threat model are trained on different datasets.
Since samples from different datasets vary in factors such as background, weather condition and camera angle, the domain gap would degrade performance.
Secondly, the proprietary model's internal workings are out of reach because the adversary can only access it through a black-box \acrshort{api}.
Outputs of intermediate layers in the proprietary model are unattainable, and methods such as lateral shortcut connections~\cite{valpola2015neural} can not be applied.
Thirdly, the threat model can not be optimized simultaneously with the proprietary model since the proprietary model is fixed.
It leads to a mismatch between the objectives of the proprietary model and the threat model, \eg, the proprietary model learns representative features for facial recognition while the threat model is trained to reconstruct face images.

\section{ShuffleBits}

In spite of recent studies on binarized neural networks~\cite{rastegari2016xnor,courbariaux2016binarized} which reduce memory consumption and improve inference speed, storing weights and activations in the single-precision floating-point format is still the predominant option.
Each single-precision floating-point value can be viewed as a 32-bit binary sequence (\ie, binary32).
The IEEE 754 standard~\cite{IEEE_754} defines the procedure which converts a real number from decimal representation to binary32 format, and vice versa.

Given a single-precision floating-point value $x$, it can be represented as a finite binary sequence
\begin{equation}
(a_{i})_{i \in I},
\end{equation}
where $a_{i} \in \{0,1\}$, $I = \{1,\ldots,n\}$ and $n = 32$.

One could shuffle the original sequence according to an encryption key, and the encrypted sequence is
\begin{equation}
(b_{j})_{j \in J},
\end{equation}
where $J = \{1,\ldots,n\}$.
The encryption key is a bijective function $f \colon I \rightarrow J$, and it is an injective and surjective mapping of set $I$ to set $J$.
In addition, we have $b_{f(i)} = a_{i}$ for $i \in I$.

Similarly, the decrypted sequence is
\begin{equation}
(c_{k})_{k \in K},
\end{equation}
where $K = \{1,\ldots,n\}$.
The decryption key is another bijective function $g \colon J \rightarrow K$ which maps set $J$ to set $K$.
Furthermore, we have $c_{g(j)} = b_{j}$ for $j \in J$.

Since $f$ is a bijection, it has an inverse function obtained by swapping the inputs and outputs in $f$.
Let $g$ be the inverse function of $f$, we have
\begin{equation}
a_{i} = b_{f(i)} = c_{g(f(i))} = c_{i} \text{ for } i \in I.
\end{equation}

With the correct decryption key, it is apparent that the decrypted sequence is identical to the original sequence.
Finally, the encrypted sequence and the decrypted sequence can be converted to decimal representation.

\begin{figure}[t]
\begin{center}
\includegraphics[width=1.0\linewidth]{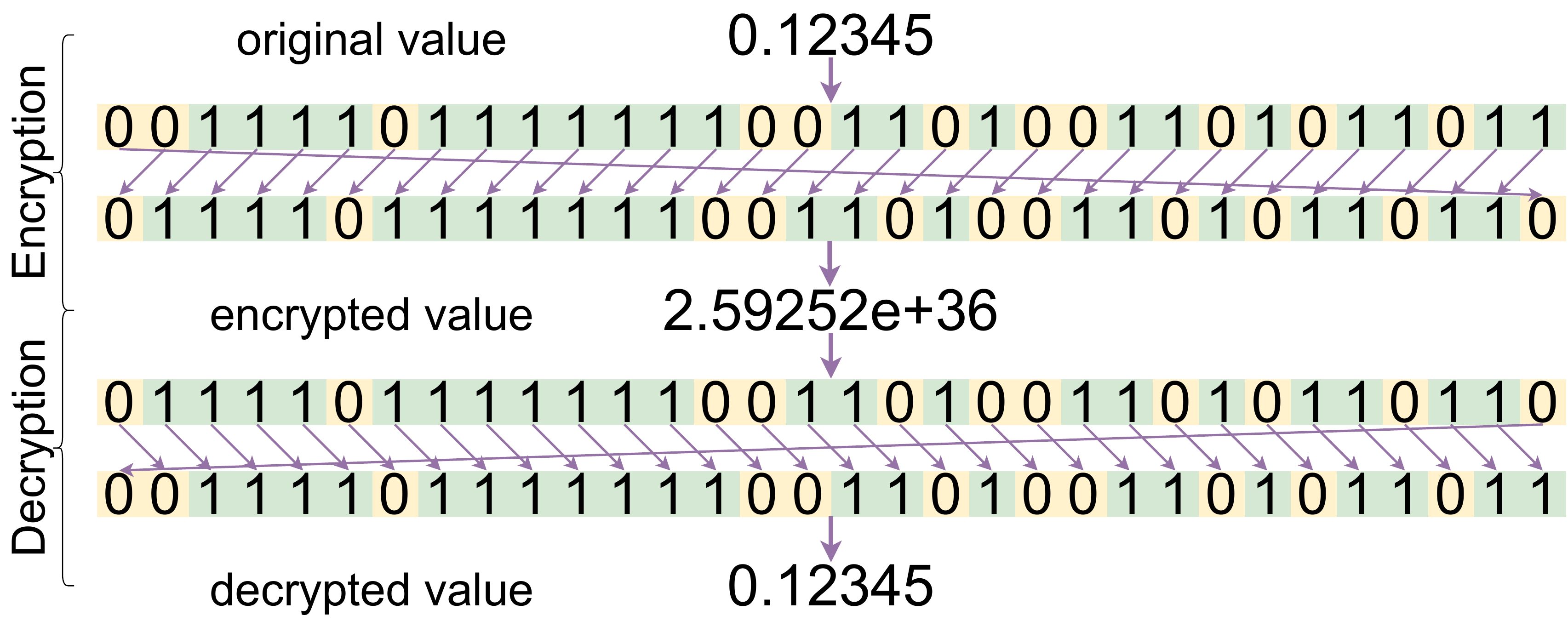}
\end{center}
\caption{
A specific case of ShuffleBits:
a left rotation operation is applied in the encryption process,
and a right rotation operation is applied in the decryption process.
}
\label{figure:shuffle_bits}
\end{figure}

Figure~\ref{figure:shuffle_bits} provides a step-by-step explanation of the proposed method using specific encryption and decryption keys.
The original value's binary sequence is shuffled according to the encryption key, and the encrypted sequence corresponds to the encrypted value.
By contrast, modifications are reverted so that the decrypted value is the same as the original value.

In the event of a brute-force attack, the adversary must systematically enumerate all possible decryption keys and check each of them.
It is computationally infeasible to conduct exhaustive key search for three reasons.
Firstly, there are $32! \approx 2.63\mathrm{e}{+35}$ unique keys, thus the number of candidate keys is large.
Secondly, it is not straightforward to validate whether the decrypted values are correct or not.
Thirdly, bit shuffling can be deployed using the \acrlong{otp} scheme, and the decryption keys in each request would be different.

\section{Experiments}

\subsection{Background}

\noindent\textbf{Domain.}
We conduct experiments in the domain of person re-identification, in which the objective is to retrieve a person of interest across multiple cameras~\cite{ye2020deep}.

\noindent\textbf{Datasets.}
We select the following datasets that are widely used:
Market-1501~\cite{zheng2015scalable}, DukeMTMC-reID~\cite{ristani2016performance} and MSMT17~\cite{wei2018person}.
In each dataset, there are three partitions, namely, training set, query set, and gallery set.
The latter two sets are merged as the test set.
Throughout this study, we use MSMT17 as the proprietary dataset, while the local dataset is either Market-1501 or DukeMTMC-reID.

\noindent\textbf{Models.}
The FastReID repository provides a unified instance re-identification library, along with a set of pre-trained models~\cite{he2020fastreid}.
We include three top-performing methods which are built using the ResNet50~\cite{he2016deep} backbone, \ie, BoT~\cite{luo2019bag}, AGW~\cite{ye2020deep} and SBS~\cite{he2020fastreid}.

\begin{figure*}[t]
\centering
\begin{tabular}{@{}cc@{}}
\subfigure[][Scores on the test set in Market-1501.]{
\includegraphics[width=0.475\linewidth]{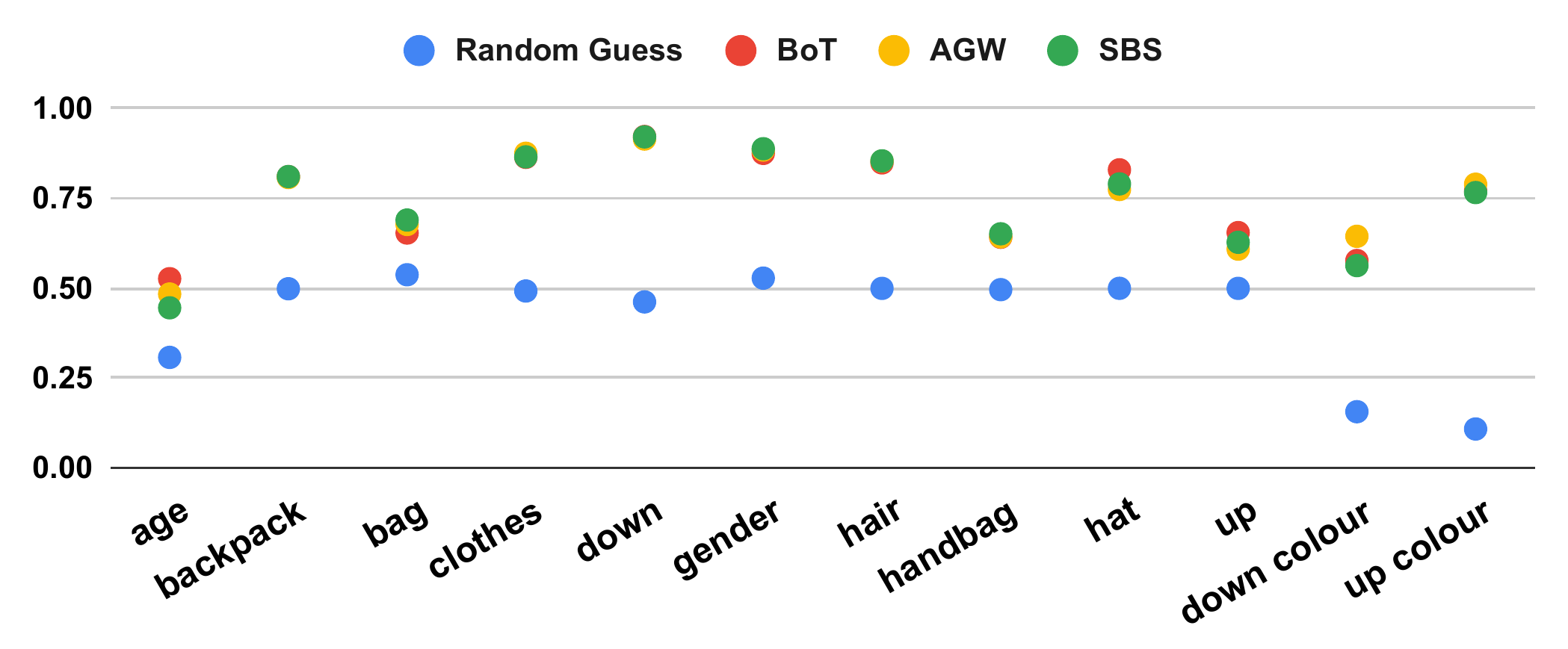}
} &
\subfigure[][Scores on the test set in DukeMTMC-reID.]{
\includegraphics[width=0.475\linewidth]{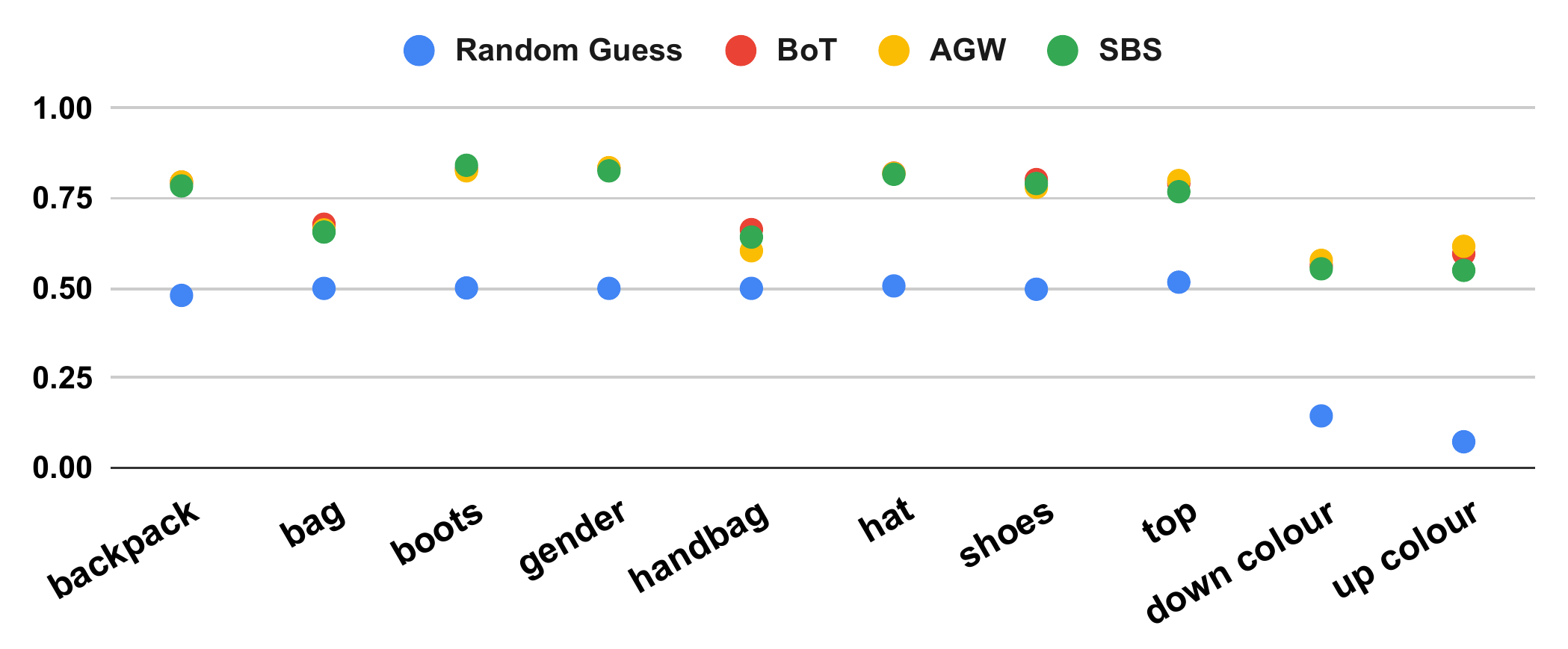}
}
\end{tabular}
\caption{
The balanced accuracies of each auxiliary attribute.
}
\label{figure:attributes}
\end{figure*}

\subsection{Recognizing auxiliary attributes}
\label{section:recognizing_auxiliary_attributes}

\noindent\textbf{Model.}
For each auxiliary attribute, batch normalization~\cite{ioffe2015batch} and dense layers are stacked to obtain the probabilities of each class.
Similar to the proprietary models~\cite{luo2019bag,ye2020deep,he2020fastreid} that classify person identities, we use only one batch normalization layer and one dense layer.
The dimensionality of the output space in the dense layer equals the number of classes.
Opting for this relatively simple architecture gives the best results in our experiments.

\noindent\textbf{Loss function.}
Similar to conventional classification models~\cite{he2016deep}, the cross-entropy loss~\cite{zhang2018generalized} is utilized on the outputs of dense layers.
Given an imbalanced dataset with unequal distribution of classes, classifiers would be biased in favor of the dominant classes.
To address this issue, we assign a scalar value to each class during training so that more attention is paid to the under-represented classes~\cite{panchapagesan2016multi}.
The class weights are inversely proportional to the count number of occurrences of each class.

\noindent\textbf{Evaluation metric.}
The accuracy score measures the percentage of samples in which the predicted label matches the corresponding ground truth.
However, it may give inflated performance estimates on imbalanced datasets.
Thus, we adopt balanced accuracy~\cite{mosley2013balanced} which is a better option, and it is defined as the average of recall calculated on each class.

\noindent\textbf{Implementation.}
The batch size is set to $128$, and the number of epochs is limited to $100$.
The Adam~\cite{kingma2014adam} optimizer is utilized in training the model.
The learning rate is fixed to $5\mathrm{e}{-5}$ in the first $50$ epochs, and it decreases by a factor of five in the remaining epochs.
The mean of balanced accuracies of all labels is monitored so that the optimal model can be identified.

\noindent\textbf{Analysis.}
We leverage the auxiliary attributes in~\cite{lin2019improving}.
These annotations provide detailed descriptions of pedestrians.
Multiple labels are present while each label corresponds to a binary or multi-class classification problem.
Figure~\ref{figure:attributes} visualizes the balanced accuracies of each auxiliary attribute in two local datasets.
Using feature vectors extracted by proprietary models yields significantly more accurate predictions than guessing randomly, and it gives coarse-grained estimations of user data.

\begin{figure*}[t]
\centering
\begin{tabular}{@{}cc@{}}
\subfigure[][Samples from the test set in Market-1501.]{
\includegraphics[width=0.475\linewidth]{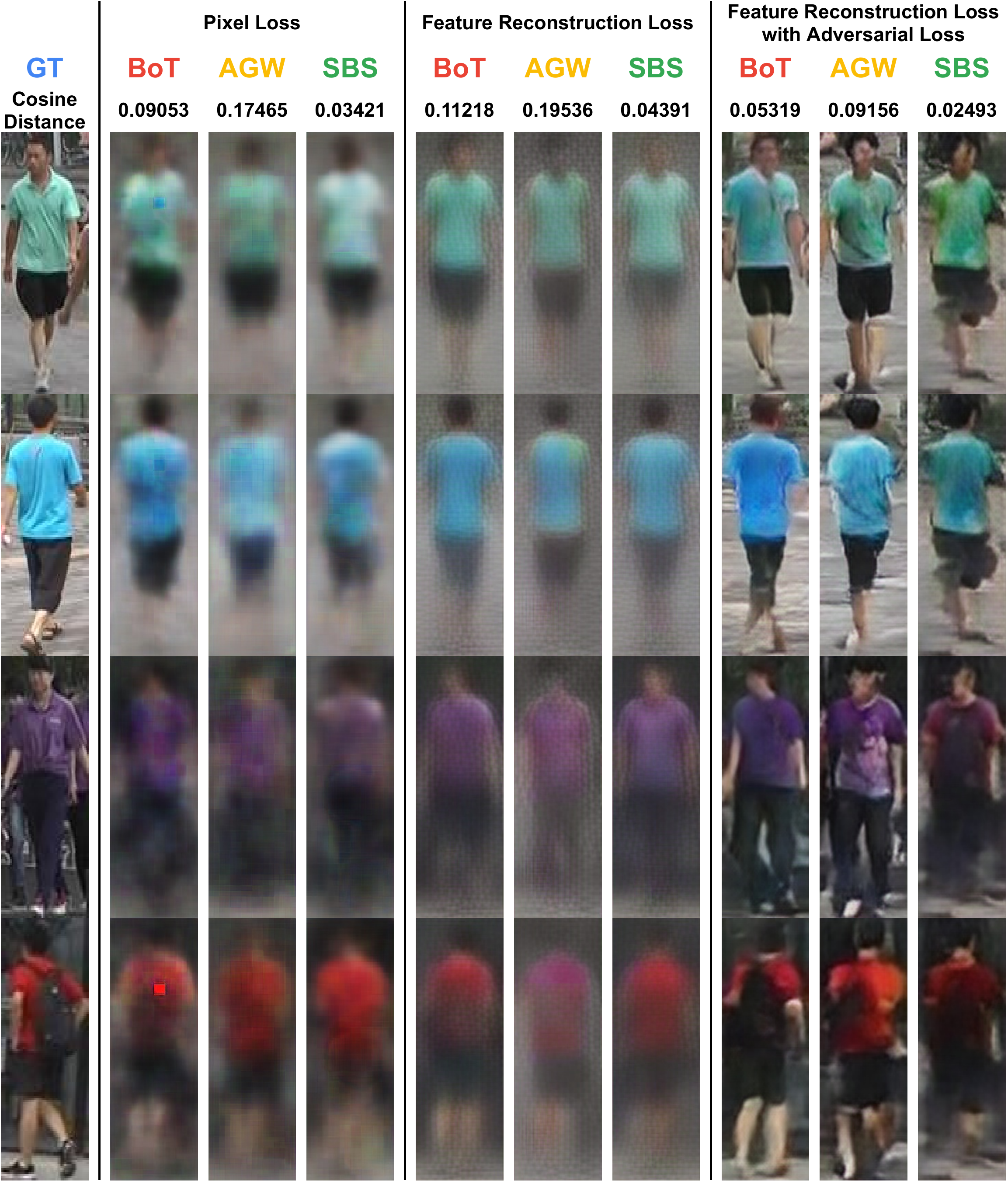}
} &
\subfigure[][Samples from the test set in DukeMTMC-reID.]{
\includegraphics[width=0.475\linewidth]{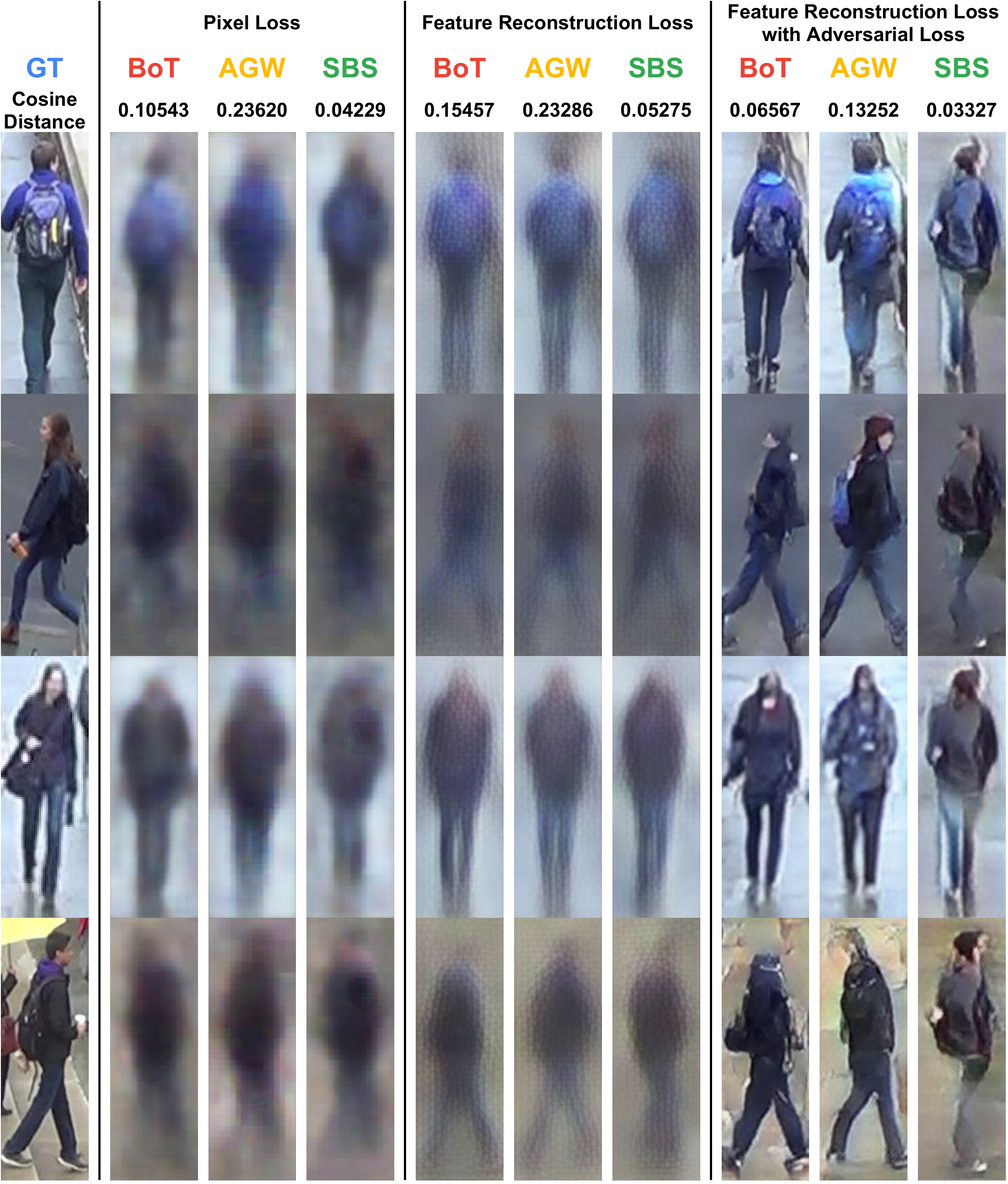}
}
\end{tabular}
\caption{
Comparison of the reconstructed images using different proprietary models, local datasets, and loss functions.
}
\label{figure:reconstructions}
\end{figure*}

\subsection{Reconstructing user data}
\label{section:reconstructing_user_data}

\noindent\textbf{Model.}
Two sub-models are involved in reconstructing user data, \ie, a generator and a discriminator.
We instantiate models with similar architectures to the BigGAN~\cite{brock2018large} work.
On the one hand, the generator maps feature vectors to images.
At the start, a dense layer with $256$ units transforms the feature vectors into a low-dimensional space, while another dense layer and a reshaping operation generate the smallest feature maps.
Subsequently, five upsampling residual blocks increase the spatial dimensionality to the target resolution (\ie, $128$×$384$).
The last convolutional layer reduces the number of channels to $3$, and the resulting predictions are in the RGB color space.
On the other hand, the discriminator classifies whether the images are original or synthetic.
Five downsampling residual blocks decrease the spatial dimensionality, and a global average pooling layer generates flattened feature vectors of the images.
Complemented with the feature vectors extracted by the proprietary model, a dense layer generates the estimations based on the concatenated feature vectors.
More details regarding the architectural layout of residual blocks can be found in the appendix of~\cite{brock2018large}.

\noindent\textbf{Loss function.}
Different loss functions are utilized for updating the generator and discriminator.
The generator can be optimized with a weighted sum of the following loss functions:
(1) The pixel loss~\cite{johnson2016perceptual} calculates the \acrlong{mse} between the ground truth images and the reconstructed images;
(2) Given a pre-trained model, one may extract an intermediate layer's outputs as feature maps.
The feature reconstruction loss~\cite{johnson2016perceptual} refers to the \acrlong{mse} between the feature maps of the ground truth images and the reconstructed images.
More specifically, we use the outputs of layer "conv2\_block3\_out" in a ResNet50~\cite{he2016deep} model that is pre-trained on the ImageNet~\cite{deng2009imagenet} dataset;
(3) The adversarial loss~\cite{goodfellow2014generative} measures how well the generator can fool the discriminator when feeding the outputs of the generator to the discriminator.
By contrast, the discriminator is optimized using the \acrlong{mse} loss that is proposed in~\cite{mao2017least}.
Compared with the cross-entropy loss~\cite{zhang2018generalized}, it suppresses the vanishing gradients problem and stabilizes the learning process.

\newpage
\noindent\textbf{Evaluation metric.}
For each reconstructed image, the ground truth image is available for reference.
Instead of comparing these images pixel by pixel, we extract the reconstructed images' feature vectors using the same proprietary model and calculate the cosine distance between feature vector pairs.
It follows the same principle as the feature reconstruction loss~\cite{johnson2016perceptual}, while there exist differences in the underlying model and the distance metric.
The cosine distance metric is widely used when comparing two feature vectors extracted by person re-identification models.
The cosine distance scores range from $0$ to $1$, and the minimum value $0$ is obtained when the angle between feature vectors is $0$.
The aforementioned evaluation metric provides quantitative performance measures for studying the effects of loss functions.

\noindent\textbf{Implementation.}
The batch size is set to $64$, and the number of iterations is limited to $60,000$.
The Adam~\cite{kingma2014adam} optimizer is utilized in training both the generator and discriminator, with the learning rate fixed to $1\mathrm{e}{-4}$.
The training procedure within one batch can be separated into multiple consecutive steps.
Firstly, the generator produces the reconstructed images based on the pre-computed feature vectors.
Secondly, the discriminator is updated by feeding the ground truth or reconstructed images alongside the corresponding labels.
Thirdly, the generator is updated while keeping the parameters of the discriminator frozen.

\noindent\textbf{Analysis.}
Generating perfect reconstructions merely from feature vectors is a difficult task.
Otherwise, the algorithm can be treated as a promising solution for neural image compression.
As discussed in Section~\ref{section:attack_scenarios}, there are three constraints that pose significant challenges, namely, the domain gap in datasets, the nature of having only black-box access, and the mismatch between optimization objectives.
Figure~\ref{figure:reconstructions} illustrates randomly chosen reconstructed images under various settings.
Those experiments differ in terms of the proprietary model, the local dataset, and the loss function.
In addition, the mean of cosine distance scores is calculated on the corresponding test set.
Three observations can be drawn:
\begin{itemize}
\itemsep0em
\item
Using the pixel loss gives inherently blurry predictions for the reason that the Euclidean distance is minimized by averaging all plausible outputs~\cite{isola2017image}.
\item
Switching to the feature reconstruction loss sharpens the images, while noticeable checkerboard artifacts are present.
\item
The checkerboard artifacts can be suppressed significantly by adding the adversarial loss, and the reconstructed images share strong similarities with the ground truth images.
Furthermore, combining the feature reconstruction loss with the adversarial loss results in the lowest cosine distance score.
\end{itemize}

\subsection{Importance of encrypting deep features}

Results in Section~\ref{section:recognizing_auxiliary_attributes} and~\ref{section:reconstructing_user_data} demonstrate that an adversary could successfully infer sensitive information even under severe constraints.
In particular, it is still feasible to recognize auxiliary attributes with decent accuracy and reconstruct user data that are recognizable.

Given a machine learning model in production, it is of great importance to adopt an encryption method.
Under the condition that an encryption method is utilized, one has to include an encryption key in each request, and feature vectors in the corresponding response are encrypted (see Figure~\ref{figure:attack_scenarios}).
Since the decryption key is kept on the client side, the original values can be recovered without changes.
The adversary could still train threat models on original feature vectors.
However, the decryption key required to decrypt feature vectors of user data is unknown, while the threat models would not generate meaningful predictions on encrypted feature vectors.
In addition, training threat models directly on encrypted feature vectors is not an option because the users and the adversary are using different encryption keys.

\subsection{ShuffleBits vs traditional encryption methods}

The ShuffleBits method differs from traditional encryption methods in two aspects.
On the one hand, computations in ShuffleBits can be translated into operations on tensors.
Incorporating ShuffleBits into an existing neural network is straightforward, and it can be implemented as a plug-and-play module without extra dependencies.
On the other hand, a model with ShuffleBits would generate encrypted deep features directly.
It eliminates the risk of transferring unencrypted data from \acrshort{gpu} to \acrshort{cpu}.

The \acrfull{aes}~\cite{daemen1999aes} method is widely accepted as the de facto standard for symmetric-key algorithms.
Since the security of ShuffleBits is yet to be validated, a natural extension would be to develop dedicated attacks against ShuffleBits from the perspective of cryptography.
In the work of InstaHide~\cite{huang2020instahide}, a method is introduced to encrypt training images in a federated learning scenario.
However, it is later found to be insecure in~\cite{carlini2021private}.
To address the potential vulnerabilities in ShuffleBits, we introduce two workarounds.
On the one hand, the \acrlong{otp} scheme can be utilized.
In each request, different encryption/decryption keys are used.
It is unlikely to recover the original feature vectors by observing a few instances.
On the other hand, ShuffleBits can be seamlessly applied alongside traditional encryption methods.
Such cascade encryption pipeline leads to better security, as an adversary has to break all the encryption algorithms to obtain useful information.

\section{Conclusion}

This study emphasizes the importance of encrypting deep features when deploying a machine learning model in production.
On the one hand, we adopt an experimental setting with only two assumptions, and it is more practical than previous works.
Two attack scenarios have been proposed to reverse state-of-the-art person re-identification models.
Results show that an adversary could recognize auxiliary attributes and reconstruct user data, thus breaching user privacy.
On the other hand, adopting an encryption method when transferring and storing deep features would prevent model inversion attacks.
By performing manipulations on the binary sequence of each floating-point number, we introduce the ShuffleBits method, and it can be implemented as a plug-and-play module inside neural networks.

{\small
\bibliographystyle{ieee_fullname}
\bibliography{nixingyang_references}

\begin{thebibliography}{10}\itemsep=-1pt

\bibitem{IEEE_754}
{IEEE Standard for Floating-Point Arithmetic}.
\newblock {\em IEEE Std 754-2019 (Revision of IEEE 754-2008)}, pages 1--84,
  2019.

\bibitem{brock2018large}
Andrew Brock, Jeff Donahue, and Karen Simonyan.
\newblock {Large scale gan training for high fidelity natural image synthesis}.
\newblock {\em arXiv preprint arXiv:1809.11096}, 2018.

\bibitem{carlini2021private}
Nicholas Carlini, Samuel Deng, Sanjam Garg, Somesh Jha, Saeed Mahloujifar,
  Mohammad Mahmoody, Shuang Song, Abhradeep Thakurta, and Florian Tramer.
\newblock {Is Private Learning Possible with Instance Encoding?}, 2021.

\bibitem{courbariaux2016binarized}
Matthieu Courbariaux, Itay Hubara, Daniel Soudry, Ran El-Yaniv, and Yoshua
  Bengio.
\newblock {Binarized neural networks: Training deep neural networks with
  weights and activations constrained to+ 1 or-1}.
\newblock {\em arXiv preprint arXiv:1602.02830}, 2016.

\bibitem{daemen1999aes}
Joan Daemen and Vincent Rijmen.
\newblock {AES proposal: Rijndael}.
\newblock 1999.

\bibitem{deng2009imagenet}
Jia Deng, Wei Dong, Richard Socher, Li-Jia Li, Kai Li, and Li Fei-Fei.
\newblock {ImageNet: A large-scale hierarchical image database}.
\newblock In {\em IEEE Conference on Computer Vision and Pattern Recognition},
  pages 248--255. Ieee, 2009.

\bibitem{dong2018boosting}
Yinpeng Dong, Fangzhou Liao, Tianyu Pang, Hang Su, Jun Zhu, Xiaolin Hu, and
  Jianguo Li.
\newblock {Boosting adversarial attacks with momentum}.
\newblock In {\em Proceedings of the IEEE conference on computer vision and
  pattern recognition}, pages 9185--9193, 2018.

\bibitem{dosovitskiy2016generating}
Alexey Dosovitskiy and Thomas Brox.
\newblock {Generating images with perceptual similarity metrics based on deep
  networks}.
\newblock {\em arXiv preprint arXiv:1602.02644}, 2016.

\bibitem{dosovitskiy2016inverting}
Alexey Dosovitskiy and Thomas Brox.
\newblock {Inverting visual representations with convolutional networks}.
\newblock In {\em Proceedings of the IEEE conference on computer vision and
  pattern recognition}, pages 4829--4837, 2016.

\bibitem{fredrikson2015model}
Matt Fredrikson, Somesh Jha, and Thomas Ristenpart.
\newblock {Model inversion attacks that exploit confidence information and
  basic countermeasures}.
\newblock In {\em Proceedings of the 22nd ACM SIGSAC Conference on Computer and
  Communications Security}, pages 1322--1333, 2015.

\bibitem{fredrikson2014privacy}
Matthew Fredrikson, Eric Lantz, Somesh Jha, Simon Lin, David Page, and Thomas
  Ristenpart.
\newblock {Privacy in pharmacogenetics: An end-to-end case study of
  personalized warfarin dosing}.
\newblock In {\em 23rd USENIX Security Symposium (USENIX Security 14)}, pages
  17--32, 2014.

\bibitem{goodfellow2014generative}
Ian Goodfellow, Jean Pouget-Abadie, Mehdi Mirza, Bing Xu, David Warde-Farley,
  Sherjil Ozair, Aaron Courville, and Yoshua Bengio.
\newblock {Generative adversarial nets}.
\newblock In {\em Advances in Neural Information Processing Systems}, pages
  2672--2680, 2014.

\bibitem{he2016deep}
Kaiming He, Xiangyu Zhang, Shaoqing Ren, and Jian Sun.
\newblock {Deep residual learning for image recognition}.
\newblock In {\em Proceedings of the IEEE Conference on Computer Vision and
  Pattern Recognition}, pages 770--778, 2016.

\bibitem{he2020fastreid}
Lingxiao He, Xingyu Liao, Wu Liu, Xinchen Liu, Peng Cheng, and Tao Mei.
\newblock {Fastreid: A pytorch toolbox for general instance re-identification}.
\newblock {\em arXiv preprint arXiv:2006.02631}, 6(7):8, 2020.

\bibitem{he2017adversarial}
Warren He, James Wei, Xinyun Chen, Nicholas Carlini, and Dawn Song.
\newblock {Adversarial example defense: Ensembles of weak defenses are not
  strong}.
\newblock In {\em 11th USENIX Workshop on Offensive Technologies (WOOT 17)},
  2017.

\bibitem{huang2020instahide}
Yangsibo Huang, Zhao Song, Kai Li, and Sanjeev Arora.
\newblock {Instahide: Instance-hiding schemes for private distributed
  learning}.
\newblock In {\em International Conference on Machine Learning}, pages
  4507--4518. PMLR, 2020.

\bibitem{ioffe2015batch}
Sergey Ioffe and Christian Szegedy.
\newblock {Batch normalization: Accelerating deep network training by reducing
  internal covariate shift}.
\newblock {\em arXiv preprint arXiv:1502.03167}, 2015.

\bibitem{isola2017image}
Phillip Isola, Jun-Yan Zhu, Tinghui Zhou, and Alexei~A Efros.
\newblock {Image-to-image translation with conditional adversarial networks}.
\newblock In {\em Proceedings of the IEEE conference on computer vision and
  pattern recognition}, pages 1125--1134, 2017.

\bibitem{johnson2016perceptual}
Justin Johnson, Alexandre Alahi, and Li Fei-Fei.
\newblock {Perceptual losses for real-time style transfer and
  super-resolution}.
\newblock In {\em European conference on computer vision}, pages 694--711.
  Springer, 2016.

\bibitem{juuti2019prada}
Mika Juuti, Sebastian Szyller, Samuel Marchal, and N Asokan.
\newblock {PRADA: protecting against DNN model stealing attacks}.
\newblock In {\em 2019 IEEE European Symposium on Security and Privacy
  (EuroS{\&}P)}, pages 512--527. IEEE, 2019.

\bibitem{kingma2014adam}
Diederik~P Kingma and Jimmy Ba.
\newblock {Adam: A method for stochastic optimization}.
\newblock {\em arXiv preprint arXiv:1412.6980}, 2014.

\bibitem{krishna2019thieves}
Kalpesh Krishna, Gaurav~Singh Tomar, Ankur~P Parikh, Nicolas Papernot, and
  Mohit Iyyer.
\newblock {Thieves on sesame street! model extraction of bert-based apis}.
\newblock {\em arXiv preprint arXiv:1910.12366}, 2019.

\bibitem{kurakin2016adversarial}
Alexey Kurakin, Ian Goodfellow, Samy Bengio, and {others}.
\newblock {Adversarial examples in the physical world}, 2016.

\bibitem{lin2019improving}
Yutian Lin, Liang Zheng, Zhedong Zheng, Yu Wu, Zhilan Hu, Chenggang Yan, and Yi
  Yang.
\newblock {Improving person re-identification by attribute and identity
  learning}.
\newblock {\em Pattern Recognition}, 2019.

\bibitem{long2017towards}
Yunhui Long, Vincent Bindschaedler, and Carl~A Gunter.
\newblock {Towards measuring membership privacy}.
\newblock {\em arXiv preprint arXiv:1712.09136}, 2017.

\bibitem{luo2019bag}
Hao Luo, Youzhi Gu, Xingyu Liao, Shenqi Lai, and Wei Jiang.
\newblock {Bag of tricks and a strong baseline for deep person
  re-identification}.
\newblock In {\em Proceedings of the IEEE/CVF Conference on Computer Vision and
  Pattern Recognition Workshops}, 2019.

\bibitem{mahendran2016visualizing}
Aravindh Mahendran and Andrea Vedaldi.
\newblock {Visualizing deep convolutional neural networks using natural
  pre-images}.
\newblock {\em International Journal of Computer Vision}, 120(3):233--255,
  2016.

\bibitem{mao2017least}
Xudong Mao, Qing Li, Haoran Xie, Raymond Y~K Lau, Zhen Wang, and Stephen
  Paul~Smolley.
\newblock {Least squares generative adversarial networks}.
\newblock In {\em Proceedings of the IEEE international conference on computer
  vision}, pages 2794--2802, 2017.

\bibitem{mosley2013balanced}
Lawrence Mosley.
\newblock {A balanced approach to the multi-class imbalance problem}.
\newblock 2013.

\bibitem{oord2016conditional}
Aaron van~den Oord, Nal Kalchbrenner, Oriol Vinyals, Lasse Espeholt, Alex
  Graves, and Koray Kavukcuoglu.
\newblock {Conditional image generation with pixelcnn decoders}.
\newblock {\em arXiv preprint arXiv:1606.05328}, 2016.

\bibitem{orekondy2019knockoff}
Tribhuvanesh Orekondy, Bernt Schiele, and Mario Fritz.
\newblock {Knockoff nets: Stealing functionality of black-box models}.
\newblock In {\em Proceedings of the IEEE/CVF Conference on Computer Vision and
  Pattern Recognition}, pages 4954--4963, 2019.

\bibitem{panchapagesan2016multi}
Sankaran Panchapagesan, Ming Sun, Aparna Khare, Spyros Matsoukas, Arindam
  Mandal, Björn Hoffmeister, and Shiv Vitaladevuni.
\newblock {Multi-task learning and weighted cross-entropy for DNN-based keyword
  spotting.}
\newblock In {\em Interspeech}, volume~9, pages 760--764, 2016.

\bibitem{rastegari2016xnor}
Mohammad Rastegari, Vicente Ordonez, Joseph Redmon, and Ali Farhadi.
\newblock {Xnor-net: Imagenet classification using binary convolutional neural
  networks}.
\newblock In {\em European conference on computer vision}, pages 525--542.
  Springer, 2016.

\bibitem{ristani2016performance}
Ergys Ristani, Francesco Solera, Roger Zou, Rita Cucchiara, and Carlo Tomasi.
\newblock {Performance measures and a data set for multi-target, multi-camera
  tracking}.
\newblock In {\em European Conference on Computer Vision}, pages 17--35.
  Springer, 2016.

\bibitem{shokri2017membership}
Reza Shokri, Marco Stronati, Congzheng Song, and Vitaly Shmatikov.
\newblock {Membership inference attacks against machine learning models}.
\newblock In {\em 2017 IEEE Symposium on Security and Privacy (SP)}, pages
  3--18. IEEE, 2017.

\bibitem{su2019one}
Jiawei Su, Danilo~Vasconcellos Vargas, and Kouichi Sakurai.
\newblock {One pixel attack for fooling deep neural networks}.
\newblock {\em IEEE Transactions on Evolutionary Computation}, 23(5):828--841,
  2019.

\bibitem{tramer2016stealing}
Florian Tram{\`{e}}r, Fan Zhang, Ari Juels, Michael~K Reiter, and Thomas
  Ristenpart.
\newblock {Stealing Machine Learning Models via Prediction APIs}.
\newblock In {\em Proceedings of the 25th USENIX Conference on Security
  Symposium}, pages 601--618, 2016.

\bibitem{valpola2015neural}
Harri Valpola.
\newblock {From neural PCA to deep unsupervised learning}.
\newblock In {\em Advances in independent component analysis and learning
  machines}, pages 143--171. Elsevier, 2015.

\bibitem{wei2018person}
Longhui Wei, Shiliang Zhang, Wen Gao, and Qi Tian.
\newblock {Person transfer gan to bridge domain gap for person
  re-identification}.
\newblock In {\em Proceedings of the IEEE Conference on Computer Vision and
  Pattern Recognition}, pages 79--88, 2018.

\bibitem{wu2016methodology}
Xi Wu, Matthew Fredrikson, Somesh Jha, and Jeffrey~F Naughton.
\newblock {A methodology for formalizing model-inversion attacks}.
\newblock In {\em 2016 IEEE 29th Computer Security Foundations Symposium
  (CSF)}, pages 355--370. IEEE, 2016.

\bibitem{yang2019adversarial}
Ziqi Yang, Ee-Chien Chang, and Zhenkai Liang.
\newblock {Adversarial neural network inversion via auxiliary knowledge
  alignment}.
\newblock {\em arXiv preprint arXiv:1902.08552}, 2019.

\bibitem{ye2020deep}
Mang Ye, Jianbing Shen, Gaojie Lin, Tao Xiang, Ling Shao, and Steven C~H Hoi.
\newblock {Deep learning for person re-identification: A survey and outlook}.
\newblock {\em arXiv preprint arXiv:2001.04193}, 2020.

\bibitem{yeom2018privacy}
Samuel Yeom, Irene Giacomelli, Matt Fredrikson, and Somesh Jha.
\newblock {Privacy risk in machine learning: Analyzing the connection to
  overfitting}.
\newblock In {\em 2018 IEEE 31st Computer Security Foundations Symposium
  (CSF)}, pages 268--282. IEEE, 2018.

\bibitem{yin2020dreaming}
Hongxu Yin, Pavlo Molchanov, Jose~M Alvarez, Zhizhong Li, Arun Mallya, Derek
  Hoiem, Niraj~K Jha, and Jan Kautz.
\newblock {Dreaming to distill: Data-free knowledge transfer via
  deepinversion}.
\newblock In {\em Proceedings of the IEEE/CVF Conference on Computer Vision and
  Pattern Recognition}, pages 8715--8724, 2020.

\bibitem{zhang2020secret}
Yuheng Zhang, Ruoxi Jia, Hengzhi Pei, Wenxiao Wang, Bo Li, and Dawn Song.
\newblock {The secret revealer: Generative model-inversion attacks against deep
  neural networks}.
\newblock In {\em Proceedings of the IEEE/CVF Conference on Computer Vision and
  Pattern Recognition}, pages 253--261, 2020.

\bibitem{zhang2018generalized}
Zhilu Zhang and Mert~R Sabuncu.
\newblock {Generalized cross entropy loss for training deep neural networks
  with noisy labels}.
\newblock {\em arXiv preprint arXiv:1805.07836}, 2018.

\bibitem{zheng2015scalable}
Liang Zheng, Liyue Shen, Lu Tian, Shengjin Wang, Jingdong Wang, and Qi Tian.
\newblock {Scalable person re-identification: A benchmark}.
\newblock In {\em Proceedings of the IEEE International Conference on Computer
  Vision}, pages 1116--1124, 2015.

\bibitem{zhong2017random}
Zhun Zhong, Liang Zheng, Guoliang Kang, Shaozi Li, and Yi Yang.
\newblock {Random erasing data augmentation}.
\newblock {\em arXiv preprint arXiv:1708.04896}, 2017.

\end{thebibliography}
}

\end{document}